\pdfoutput=1
\documentclass[11pt]{article}

\usepackage[final]{acl}

\usepackage{times}
\usepackage{latexsym}

\usepackage{caption}
\usepackage{algorithm}
\usepackage{algpseudocode}
\usepackage{microtype}
\usepackage{graphicx}
\usepackage{booktabs}
\usepackage{hyperref}
\usepackage{xcolor}         % colors
\usepackage{colortbl}

\usepackage{amsmath}
\usepackage{amssymb}
\usepackage{mathtools}
\usepackage{amsthm}

\usepackage[capitalize,noabbrev]{cleveref}

\usepackage{lipsum}
\usepackage{duckuments}
\usepackage{subcaption}

\usepackage{wrapfig}
\usepackage{enumitem}
\usepackage{colortbl}
\usepackage{bm}
\usepackage{float}
\usepackage{multirow}
\usepackage{blindtext}
\usepackage{makecell}
\usepackage{adjustbox}
\usepackage{soul}

\usepackage[normalem]{ulem}
\useunder{\uline}{\ul}{}
\DeclareUnicodeCharacter{221E}{\ensuremath{\infty}}
\usepackage{cleveref}
\crefformat{table}{Table~#2#1#3}
\crefformat{figure}{Figure~#2#1#3}
\crefformat{section}{Section~#2#1#3}
\crefformat{appendix}{Appendix~#2#1#3}
\crefformat{equation}{Eq.~#2(#1)#3}

\newcommand{\mname}{STAND\xspace}

\usepackage[T1]{fontenc}
\usepackage[utf8]{inputenc}

\usepackage{microtype}

\usepackage{inconsolata}

\usepackage{graphicx}

\title{Accelerated Test-Time Scaling with Model-Free Speculative Sampling}

\author{
  Woomin Song$^{1, \dagger}$,
  \,
  Saket Dingliwal$^{2}$,
  \,
  Sai Muralidhar Jayanthi$^{2}$,
  \,
  \\
  \textbf{Bhavana Ganesh}$^{3, \ddagger}$,
  \textbf{Jinwoo Shin}$^{1}$,
  \textbf{Aram Galstyan}$^{2}$,
  \textbf{Sravan Babu Bodapati}$^{2}$
  \\
  $^{1}$KAIST
  $^{2}$Amazon AGI
  $^{3}$AirSignal
}

\usepackage[symbol]{footmisc}

\begin{document}
\maketitle

\footnotetext{\ignorespaces$^{\dagger}$ Work done during an internship at Amazon.
\ignorespaces$^{\ddagger}$ Work done at Amazon.
}
% CRv2-ACTIVE
\begin{abstract}

Language models have demonstrated remarkable capabilities in reasoning tasks through test-time scaling techniques like best-of-N sampling and tree search. However, these approaches often demand substantial computational resources, creating a critical trade-off between performance and efficiency. We introduce \mname (STochastic Adaptive N-gram Drafting), a novel model-free speculative decoding approach that exploits the inherent redundancy in reasoning trajectories to achieve significant acceleration without compromising accuracy. Our analysis shows that reasoning paths frequently reuse similar reasoning patterns, enabling efficient model-free token prediction without requiring separate draft models. By introducing stochastic drafting and preserving probabilistic information through a memory-efficient logit-based N-gram module, combined with optimized Gumbel-Top-K sampling and data-driven tree construction, \mname significantly improves token acceptance rates. Extensive evaluations across multiple models and reasoning tasks (AIME-2024, GPQA-Diamond, and LiveCodeBench) demonstrate that \mname reduces inference latency by 60-65\% compared to standard autoregressive decoding while maintaining accuracy. Furthermore, \mname consistently outperforms state-of-the-art speculative decoding methods across diverse inference patterns, including single-trajectory decoding, batch decoding, and test-time tree search.
As a model-free approach, STAND can be applied to any existing language model without additional training, making it a powerful plug-and-play solution for accelerating language model reasoning.
\end{abstract}

% CRv2-ACTIVE
\section{Introduction}
\label{sec:introduction}

Test-time scaling has emerged as a prominent paradigm for enhancing the performance of language models by allocating additional computational resources during inference \citep{snell2024scaling}. This includes generating long sequences of thoughts though Large Reasoning Models (LRMs) \cite{muennighoff2025s1}, multi-sampling approaches like best-of-N sampling and majority voting that generate multiple independent outputs to select the most promising one \citep{Wang2022SelfConsistencyIC}, as well as iterative methods like tree search and sequential refinement that allow models to progressively improve their reasoning process \citep{uesato2022solving}.
While these methods demonstrate the potential for significant accuracy improvements through increased computation, they often demand substantial computational resources due to the large number of tokens that need to be generated.

Recent research has focused on reducing the high computational costs of test-time scaling and reasoning approaches \citep{sui2025stop}. Some work has explored training with length-based rewards to generate more concise outputs \citep{aggarwal2025l1, qu2025optimizing}, while other approaches use combinations of small and large models to distribute the workload efficiently \citep{liao2025reward, yang2025speculative}.
However, these efficiency-focused methods typically face a fundamental trade-off. While they reduce computational costs, they tend to sacrifice some accuracy compared to more exhaustive approaches, as using fewer samples or cutting short the exploration process often leads to lower performance.

% CRv2-ACTIVE
\begin{figure*}[t]
    \centering
\begin{subfigure}{0.329\textwidth}
\includegraphics[width=\linewidth]{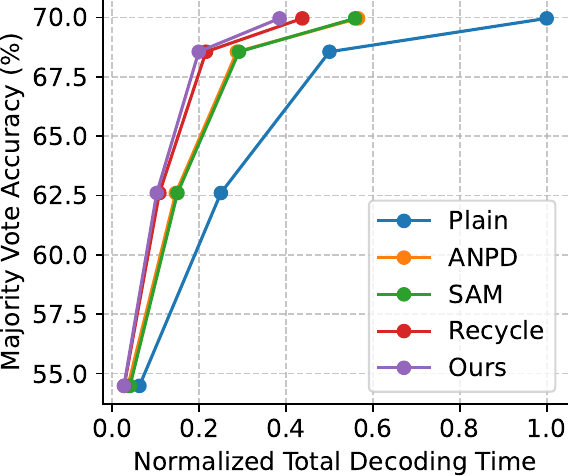}
\caption{AIME-2024}\label{fig:main-aime}
\end{subfigure}
\begin{subfigure}{0.329\textwidth}
\includegraphics[width=\linewidth]{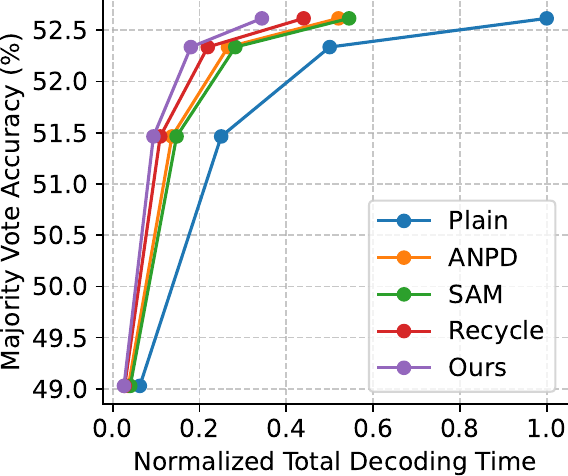}
\caption{GPQA-Diamond}\label{fig:main-gpqa}
\end{subfigure}
\begin{subfigure}{0.329\textwidth}
\includegraphics[width=\linewidth]{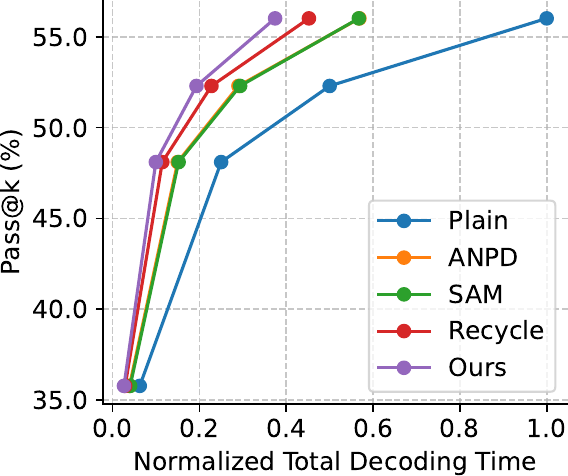}
\caption{LiveCodeBench}
\label{fig:main-lcb}
\end{subfigure}
\caption{\textbf{Scaling curve with speculative decoding.} We report the scaling curve describing how the task performance improves with respect to the total decoding time. Keeping simple auto-regressive decoding total time as 1, we also report  the scaling curves for different model-free SD methods. We report the reward-weighted majority voting accuracy for AIME-2024 and GPQA-Diamond, and pass@k for LiveCodeBench, where k is the total number of generated sequences generated at a given point.
All measurements are made on a single A100 GPU with DeepSeek-R1-Distill-Qwen-7B.}
\label{fig:main}
\end{figure*}

This raises a crucial question: How can we improve the efficiency of test-time scaling and reasoning methods without compromising their accuracy?
To address this challenge, we turn our attention to speculative decoding (SD), which offers a promising solution for lossless acceleration of language model inference.
Speculative decoding accelerates language model inference by using a smaller "draft" model to predict tokens, which are then verified by the larger target model \citep{leviathan2023fast}. With appropriate verification strategies
\citep{chen2023accelerating}, SD can speed up the auto-regressive decoding process of large language models while preserving their 
% original 
output distribution.

A key observation in LRMs is the significant repetition of token sequences across different reasoning paths. When models are performing chain-of-thought reasoning \citep{snell2024scaling} or exploring multiple solutions \citep{Wang2022SelfConsistencyIC, xie2024monte}, they frequently reuse similar expressions, logical deductions, and reasoning patterns.

This redundancy presents an opportunity for model-free speculative decoding \citep{pld, anpd}. Unlike model-based approaches that rely on neural networks as drafters \citep{li2024eagle2, cai2024medusa}, model-free methods can leverage patterns from previous generations to construct drafts. This makes them particularly well-suited for exploiting cross-trajectory information. Our experiments confirm this approach's effectiveness, demonstrating improved efficiency as the number of reasoning trajectories increases.

To fully leverage the power of model-free speculative decoding for reasoning tasks, we propose \textbf{\mname (STochastic Adaptive N-gram Drafting)}. Our approach is motivated by two key observations: First, existing model-free approaches have primarily focused on greedy decoding, leaving the potential benefits of sampling largely unexplored. 
Second, our experimental analysis demonstrates that stochastic drafting (i.e. sampling draft tokens from the draft probability distribution) significantly improves token acceptance rates.
Building on these insights, \mname introduces three key innovations: (1) a memory-efficient logit-based N-gram module that preserves probabilistic information for better stochastic drafting, (2) an optimized sampling strategy using Gumbel-Top-K for efficient token selection, and (3) a data-driven approach to draft tree construction that balances efficiency with effectiveness. Combined, these techniques significantly enhance the speculative decoding performance in the context of test-time scaling, where sampling and diverse trajectory exploration are crucial.

Our extensive evaluations demonstrate \mname's effectiveness across diverse reasoning tasks (math, science, and coding) and model scales. As highlighted in Figure \ref{fig:main}, STAND’s benefits become more pronounced as the number of reasoning trajectories increases. With best-of-16 sampling for optimal accuracy, STAND reduces inference latency by 60–65\% compared to standard autoregressive decoding while maintaining performance. Moreover, STAND outperforms state-of-the-art speculative decoding methods by 14–28\% in throughput, establishing an efficient drafting strategy for reasoning tasks. 

Furthermore, \mname consistently achieves the best throughput across multiple inference patterns, reducing the inference latency by 58\% in single-trajectory decoding, 30\% in batch decoding, and 61\% in test-time tree search, compared to standard autoregressive decoding. As a model-free speculative decoding approach, \mname accomplishes all these achievements without requiring any additional drafter model, or fine-tuning the target model, being able to be used in plug-and-play manner to any existing LRMs.
% CRv2-ACTIVE
\section{Related Works}
\label{sec:related-works}

\paragraph{Test-time scaling and efficiency.} 
Test-Time Scaling (TTS) has emerged as a prominent strategy to enhance problem-solving capabilities during inference without model retraining \cite{snell2024scaling, muennighoff2025s1, Wang2022SelfConsistencyIC, uesato2022solving, xie2024monte}. Generating long chain-of-thoughts or sampling multiple sequences have consistently showcased higher accuracy in many complex tasks like math, science and coding \cite{wei2022chain, cobbe2021training, chen2023universal}. However, the computational cost of TTS remains
a critical bottleneck for their practical use. Recent work has explored
optimizing inference using adaptive thinking lengths, cascading models of different sizes, length penalties during training, and budget-constrained decoding \cite{aggarwal2025l1, qu2025optimizing, liao2025reward, li2024escape, wan2024dynamic}, yet
the fundamental trade-off between accuracy gains
and costs persists. Our method aims to accelerate reasoning while ensuring no performance degradation.

\paragraph{Speculative decoding.}
SD have been shown to accelerate Large Language Model (LLM) inference without any loss in the accuracy \cite{leviathan2023fast}. The approach typically involves a smaller "draft" model proposing candidate token sequences for parallel verification by the larger "target" model. If the tokens align with the target model's output distribution, they are "accepted", resulting in more than one token being produced in a single forward pass of the LLM. Various compute-efficient drafting strategies have been proposed in the literature to increase the chances of acceptance. Neural draft architectures have evolved from simple, smaller LMs \cite{leviathan2023fast, choi2025mamba} to sophisticated self-drafting approaches \cite{cai2024medusa, li2024eagle, cheng2024recurrent}). 
Although the use of light-weight model-free drafters based on n-grams \cite{li2024nearest, somasundaram2024pld+,hu2024sam, oliaro2024suffixdecoding, luo2024turning, anpd, pld} has been explored previously for generic tasks, we revisit them in the context of LRMs. While these approaches limit themselves to deterministic n-gram based lookups as draft sequences, we highlight the significance of stochastic drafting with logit information of previously generated n-grams for reasoning in our proposed method. 
To further boost SD performance, tree drafting was proposed where multiple draft token predictions are organized in a tree structure, enabling efficient parallel verification through a specialized tree attention mask \cite{Miao2023SpecInfer, li2024eagle}.
Methods like Eagle-2 \cite{li2024eagle2}
even used dynamic tree layout choices for SD. Extending these existing methods, we additionally propose a computationally efficient data-driven offline tree optimization method for our lightweight model-free drafting method for LRMs. 

Other approaches in literature that tie SD with LRMs include Speculative Thinking \cite{hu2025speculative}, SpecReason \cite{pan2025specreason}, Reward-guided SD \cite{liao2025reward}. However, they do not maintain the lossless nature of SD and hence can also be used in combination with our work.  A contemporary work \cite{li2025speculative} have explored the importance of model-free n-gram based drafting for multi-sample inference, but did not showcase any practical speedup. We extend their findings with our novel model-free stochastic drafting, and showcase a comparative analysis with existing methods through our extensive experimentation. 

% CRv2-ACTIVE
\section{Motivation}
\label{sec:obs}

\subsection{N-gram overlap analysis}
\label{sec:obs-overlap}

% CRv2-ACTIVE
\begin{figure}[h]
\centering
\includegraphics[width=\linewidth]{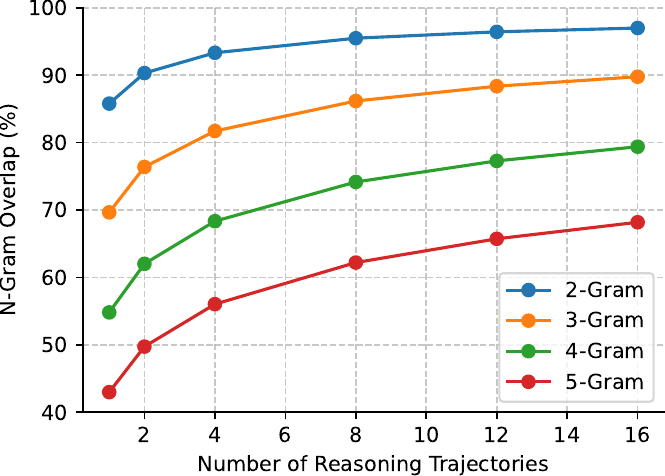}
\caption{\textbf{N-gram overlaps across reasoning trajectories.} We report the N-gram overlaps across different number of reasoning trajectories, generated by DeepSeek-R1-Distill-Qwen-7B on AIME-2024. The overlap is defined as the percentage of the N-grams that appear twice or more in the \textit{k} reasoning trajectories, counting duplicates multiple times. We observe high n-gram overlaps across reasoning paths, presenting an opportunity for faster drafting.}
\label{fig:overlap}
\end{figure}

To assess the degree of redundancy in reasoning trajectories, we conducted a comprehensive analysis of n-gram overlap patterns across multiple solutions generated by the DeepSeek-R1-Distill-Qwen-7B model on the AIME-2024 dataset. \Cref{fig:overlap} illustrates our findings, depicting the overlap rates for n-grams ranging from bigrams to 5-grams across varying numbers of reasoning trajectories.

The results reveal a substantial level of repetition in token sequences. Notably, we observed that up to 97\% of bigrams and 80\% of 4-grams recur across 16 distinct reasoning trajectories. Even when considering only two trajectories, over 90\% of bigrams are repeated. This high degree of overlap suggests a significant probability that any given n-gram generated by the model has likely appeared in a previous trajectory.

These findings present a compelling opportunity for the development of an efficient drafting strategy. By leveraging this inherent redundancy, we can implement a straightforward approach where previously generated n-grams are proposed as draft sequences, potentially leading to significant improvements in computational efficiency without compromising the chance of acceptance of the generated draft. This presents a key motivation for our proposed method \mname.

\subsection{Effectiveness of stochastic drafting}
\label{sec:obs-stochastic}

% CRv2-ACTIVE
\begin{figure}[h]
\centering
\includegraphics[width=\linewidth]{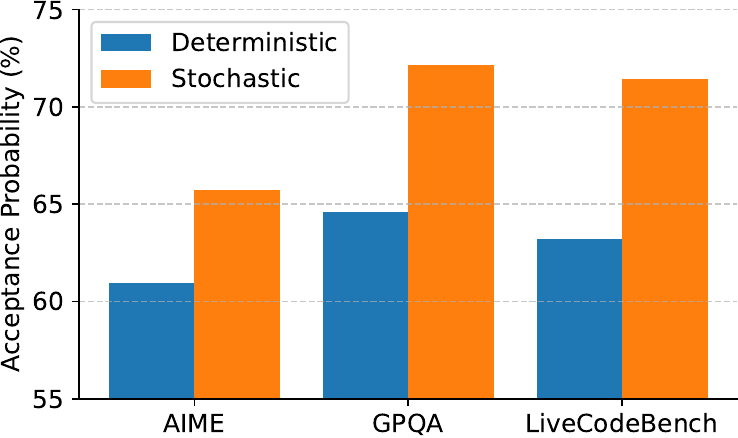}
\caption{\textbf{Deterministic vs. stochastic drafting.} We report the acceptance probability of a token, given a draft tree with depth 1 and width 3. Measurements are done using DeepSeek-R1-Distill-Qwen-7B model, and the draft tree is constructed using the N-gram module in \mname.}
\label{fig:stochastic}
\end{figure}

In contrast to traditional generation approaches that rely on greedy decoding, LRMs typically employ sampling-based generation strategies to produce multiple diverse solution trajectories, making the choice of drafting strategy particularly crucial.  
In speculative sampling \cite{chen2023accelerating}, given a target distribution $p(x)$ and draft distribution $q(x)$, the speculative sampling procedure operates by first sampling $x \sim q(x)$. The sampled token is accepted if $q(x) \leq p(x)$. Otherwise, when $q(x) > p(x)$, the token is rejected with probability $1 - \frac{p(x)}{q(x)}$ and resampled from an adjusted distribution $p'(x) = \text{norm}(\max(0, p(x) - q(x)))$. This procedure guarantees that the final output distribution matches the target distribution $p(x)$, for any drafting distribution $q(x)$.

One can choose the drafting strategy to be deterministic or stochastic. In the former, $q(x)$ is treated as a one-hot vector where $q(x_{\text{draft}}) = 1$ for the most probable token $x_{\text{draft}}$  and $q(x) = 0$ for all other $x$. For speculative sampling, this means the drafted token is accepted with $p(x_{\text{draft}})$, which can be particularly low when the target model is uncertain about its prediction. In contrast, stochastic drafting generates drafts through sampling from a probability distribution. Aligning this draft distribution with the target can significantly boost the chances of acceptance. 

In generic greedy decoding setups where this choice does not matter, existing model-free SD methods \cite{anpd, hu2024sam, pld} do not store any probability distribution with the n-gram lookup-based drafters. Eagle-2 \cite{li2024eagle2} also uses deterministic drafting for better compatibility with their dynamic tree construction logic. However, for LRMs where sampling plays a key role in generation, we showcase that this choice plays a pivotal role in acceptance probability of the draft sequence. 
As shown in Figure \ref{fig:stochastic}, this fundamental difference leads to 5\%, 7\% and 8\% higher acceptance probabilities for stochastic drafting compared to deterministic drafting  across different reasoning tasks i.e. AIME, GPQA and LiveCodeBench respectively. These experimental findings motivated us to find effective ways to compute draft model probabilities in STAND, that aligns well with the probability distributions of LRMs from which the multiple trajectories are sampled. 

% CRv2-ACTIVE
\section{\mname}
\label{sec:method}

% CRv2-ACTIVE
\begin{figure*}[t]
\centering
\includegraphics[width=\linewidth]{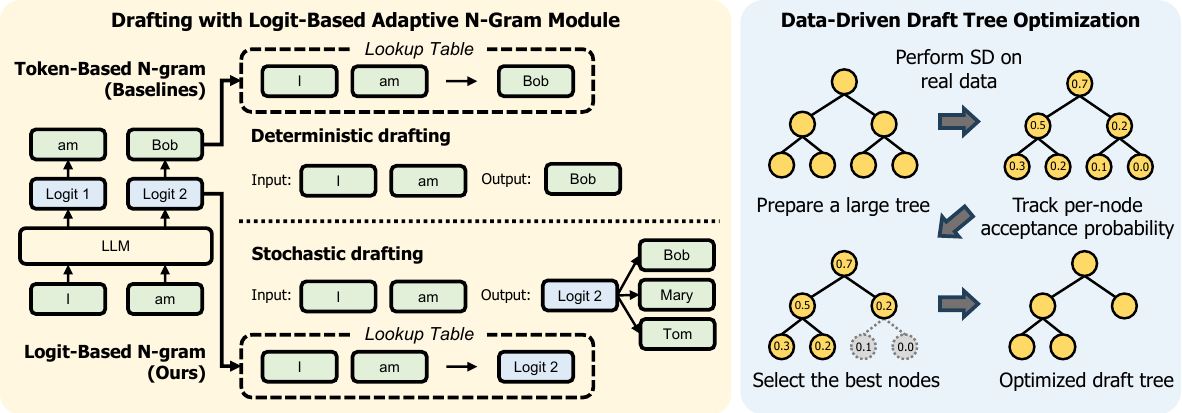}
\caption{\textbf{Overview of \mname.} 
(Left) The N-gram module stores logits instead of discrete tokens, enabling stochastic drafting. When the language model generates ``I am Bob'', we store the probability distribution over the next token rather than just the sampled token. (Right) Data-driven draft tree optimization: We start with an initial large draft tree, measure node-wise acceptance rates during speculative decoding on real data, and prune to retain the most successful paths.
}
\label{fig:method}
\end{figure*}

In this section, we present the details of \mname. In \Cref{sec:method-ngram}, we propose a memory- and compute-efficient approach to construct the logit-based N-gram module. Then in \Cref{sec:method-drafting}, we illustrate how to use the N-gram module as a drafter for stochastic sampling, together with several optimization techniques that further improve performance.

\subsection{Logit-based adaptive N-gram module}
\label{sec:method-ngram}

Traditional N-gram modules for speculative decoding typically store pairs of N-grams and their corresponding next tokens \citep{anpd}. We improve this approach by instead storing the logit distribution from which the next token is sampled. This modification preserves the rich probabilistic information of potential next tokens, enabling more sophisticated stochastic drafting strategies. While existing methods like Token Recycle partially utilize logit information by storing top-k token IDs, they discard valuable probability information that are crucial for stochastic drafting. Like previous works \citep{pld,anpd}, we maintain separate lookup tables from unigrams to 4-grams.

\paragraph{Efficient logit approximation.} To address the memory overhead associated with storing full logit distributions, particularly for models with large vocabularies, we implement a compressed representation scheme. Our approach maintains only the top-k indices and their corresponding probabilities. When encountering repeated n-grams, we merge distributions by treating non-stored indices as having zero probability and computing a weighted average: for an n-gram seen \textit{k} times previously, the existing distribution (representing the mean of \textit{k} occurrences) is weighted by \textit{k}/(\textit{k}+1) and the new distribution by 1/(\textit{k}+1). The resulting averaged distribution is then truncated to retain only the top-10 most probable tokens, ensuring constant memory usage while preserving the most relevant probability information for future speculation.

\subsection{Drafting with \mname}
\label{sec:method-drafting}

\paragraph{Stochastic tree drafting.} For each position in the draft tree, we predict the next tokens using a multi-level N-gram approach. Following previous works \citep{pld,anpd}, we search for matching N-grams in decreasing order of length, from 4-grams down to unigrams, using the first successful match. This lookup returns the top-10 candidate tokens and their corresponding probabilities from our stored distributions. Based on the number of children required at each tree node, we sample k tokens without replacement from these candidates. These sampled tokens then undergo standard speculative sampling verification to ensure draft quality.

\paragraph{Gumbel-Top-K sampling.} For efficient stochastic drafting, we replace traditional sequential sampling with a parallel sampling approach based on the Gumbel-Top-K trick \citep{gumbeltopk}. For each candidate token's log probability $\phi_i$, we add Gumbel noise to create a perturbed distribution:
$$\phi_i' = \phi_i -\log(-\log U_i),\quad U_i \sim \text{Uniform}(0,1)$$
Taking the top-k indices from these perturbed values $\phi_i'$ effectively samples k tokens without replacement in parallel, significantly reducing sampling latency compared to sequential methods.

To further optimize performance, we pre-compute and cache the Gumbel noise terms rather than generating them during drafting. This cached noise is periodically refreshed when depleted, effectively separating the sampling overhead from drafting. These optimizations further enhance the performance of our stochastic drafting approach.

\paragraph{Draft tree optimization.} Tree-based speculative decoding typically uses either dynamic trees constructed during inference or static trees built using heuristics. While dynamic trees offer context-adaptability, they add computational overhead. Conversely, static trees are computationally efficient but may underperform if constructed through heuristics alone.

We address this limitation through a data-driven approach to static tree construction. Our method begins by initializing a large tree with 625 nodes and performing speculative decoding on 30 data samples. During this process, we track which nodes are frequently part of successful speculation paths. We then select the top-80 most effective nodes and reorganize them into a compact tree structure. This empirical approach maintains the computational efficiency of static trees while ensuring the tree structure is optimized based on real-world performance data.

% CRv2-ACTIVE
\section{Experiments}
\label{sec:experiments}

This section highlights the effectiveness of \mname through extensive experiments. In \Cref{sec:experiments-multitraj}, we showcase that \mname can significantly speed up generation in multi-trajectory inference. \Cref{sec:experiments-singletraj}, we highlight that \mname can also be used in single-trajectory inference.
In \Cref{sec:experiments-batch-dvts}, we extend our evaluations to more diverse inference patterns, namely batch decoding and test-time tree search. Finally in \Cref{sec:experiments-ablations}, we perform an ablation study of the components that make \mname effective, followed by an additional analysis of the optimized tree structure.

\paragraph{Experimental setup and baselines.} Throughout the experiments, we evaluate the effectiveness of our approach on diverse tasks, including math reasoning (AIME-2024), STEM QA (GPQA-Diamond), and coding (LiveCodeBench). We perform evaluations across different model scales, including DeepSeek-R1-Distill-Qwen-7B and 14B. For all tasks, we generate maximum 32k tokens for the 7B model, and 24k tokens for the 14B model. All sampling is done with temperature 0.6. All measurements are done on a single A100 GPU, with 30 samples per task for efficient experiments.

For model-free baselines, we compare \mname against Prompt Lookup Decoding (PLD, \cite{pld}), Adaptive N-gram Parallel Decoding (ANPD, \cite{anpd}), Token Recycle (Recycle, \cite{luo2024turning}), SAM-Decoding (SAM, \cite{hu2024sam}) and a combination of SAM decoding and Token Recycle, also proposed in the SAM paper. For Static SAM (which is a component of SAM that uses a pre-constructed suffix automation from a datastore), we construct the datastore using 4k samples from the OpenThoughts-114k \citep{openthoughts} dataset. For all methods involving static draft trees, we apply our tree optimization algorithm using 30 samples from the AIME-2024 dataset unless otherwise stated.
We also compare against Eagle-2, as a representative model-based baseline. To enable long context inference, we trained all Eagle models using OpenThoughts-114k dataset, where long samples exceeding 32k tokens were truncated.

To ensure fair comparison, we conducted all experiments on a unified codebase, adapted from Spec-Bench \citep{specbench}. In this implementation, all components such as model forwarding and draft verification are shared, and the only differences lie in the drafting algorithms themselves. Furthermore, the draft length was fixed at 80 tokens for all methods, including our proposed approach and the baselines.

\paragraph{Evaluation metrics.} We adopt throughput and acceptance length as our main evaluation metrics. Throughput measures the number of tokens generated per second, computed as the total number of generated tokens divided by the total drafting time. The acceptance length quantifies the average number of tokens generated per speculation step. For both metrics, higher values indicate better performance.

\subsection{Evaluation on multi-trajectory decoding}
\label{sec:experiments-multitraj}

% CRv2-ACTIVE
\begin{table*}[!ht]
\centering
\small
\caption{\textbf{Speculative decoding performance in multi-trajectory reasoning.} We report the average throughput (T) and acceptance length (A) for multi-trajectory test-time scaling scenarios, with different number of reasoning trajectories per problem. We evaluate each model on AIME-2024 (AIME), GPQA-Diamond (GPQA), and LiveCodeBench (LCB). Best results are shown in \textbf{bold}.}

\adjustbox{width=\linewidth}{
\begin{tabular}{@{}lccccc@{\hspace{0.15in}}c@{\hspace{0.15in}}cccc@{\hspace{0.15in}}c@{\hspace{0.15in}}cccc@{}}
\toprule
                                       &                    & \multicolumn{4}{c}{4 Trajectories}                                        &  & \multicolumn{4}{c}{8 Trajectories}                                        &  & \multicolumn{4}{c}{16 Trajectories}                                       \\
\cmidrule(r){3-6} \cmidrule(lr){7-11} \cmidrule(l){12-16}
                                       &                    & AIME             & GPQA             & LCB              & Avg.             &  & AIME             & GPQA             & LCB              & Avg.             &  & AIME             & GPQA             & LCB              & Avg.             \\
\midrule
\multicolumn{16}{c}{\textit{\cellcolor[HTML]{EFEFEF}DeepSeek-R1-Distill-Qwen-7B}}                                                                                                                                                                                                                                             \\
\midrule
Plain                                  & T                  & 26.63            & 31.34            & 27.75            & 28.57            &  & 26.63            & 31.34            & 27.75            & 28.57            &  & 26.63            & 31.34            & 27.75            & 28.57            \\
\midrule
\multirow{3}{*}{Eagle-2}               & \multirow{2}{*}{T} & 29.91            & 31.69            & 27.61            & 29.74            &  & 29.91            & 31.69            & 27.61            & 29.74            &  & 29.91            & 31.69            & 27.61            & 29.74            \\
                                       &                    & (x1.12)          & (x1.01)          & (x0.99)          & (x1.04)          &  & (x1.12)          & (x1.01)          & (x0.99)          & (x1.04)          &  & (x1.12)          & (x1.01)          & (x0.99)          & (x1.04)          \\
                                       & A                  & 2.21             & 1.99             & 2.13             & 2.11             &  & 2.21             & 1.99             & 2.13             & 2.11             &  & 2.21             & 1.99             & 2.13             & 2.11             \\
\midrule
\multirow{3}{*}{PLD}                   & \multirow{2}{*}{T} & 43.93            & 50.49            & 44.01            & 46.14            &  & 44.95            & 53.04            & 45.08            & 47.69            &  & 46.60            & 53.47            & 46.02            & 48.70            \\
                                       &                    & (x1.65)          & (x1.61)          & (x1.59)          & (x1.61)          &  & (x1.69)          & (x1.69)          & (x1.62)          & (x1.67)          &  & (x1.75)          & (x1.71)          & (x1.66)          & (x1.70)          \\
                                       & A                  & 1.78             & 1.81             & 1.73             & 1.77             &  & 1.84             & 1.89             & 1.79             & 1.84             &  & 1.89             & 1.96             & 1.85             & 1.90             \\
\midrule
\multirow{3}{*}{ANPD}                  & \multirow{2}{*}{T} & 45.52            & 57.39            & 46.30            & 49.74            &  & 46.40            & 58.97            & 47.86            & 51.08            &  & 47.06            & 60.25            & 48.81            & 52.04            \\
                                       &                    & (x1.71)          & (x1.83)          & (x1.67)          & (x1.74)          &  & (x1.74)          & (x1.88)          & (x1.72)          & (x1.79)          &  & (x1.77)          & (x1.92)          & (x1.76)          & (x1.82)          \\
                                       & A                  & 1.89             & 1.97             & 1.88             & 1.91             &  & 1.92             & 2.03             & 1.91             & 1.95             &  & 1.96             & 2.11             & 1.96             & 2.01             \\
\midrule
\multirow{3}{*}{SAM}                   & \multirow{2}{*}{T} & 44.35            & 53.21            & 45.63            & 47.73            &  & 45.64            & 55.47            & 47.24            & 49.45            &  & 47.64            & 57.53            & 48.92            & 51.36            \\
                                       &                    & (x1.67)          & (x1.70)          & (x1.64)          & (x1.67)          &  & (x1.71)          & (x1.77)          & (x1.70)          & (x1.73)          &  & (x1.79)          & (x1.84)          & (x1.76)          & (x1.80)          \\
                                       & A                  & 1.81             & 1.87             & 1.85             & 1.84             &  & 1.89             & 1.96             & 1.89             & 1.91             &  & 1.97             & 2.03             & 1.95             & 1.98             \\
\midrule
\multirow{3}{*}{Recycle}               & \multirow{2}{*}{T} & 61.38            & 71.51            & 60.62            & 64.50            &  & 61.70            & 71.55            & 60.93            & 64.73            &  & 60.86            & 71.23            & 61.36            & 64.48            \\
                                       &                    & (x2.30)          & (x2.28)          & (x2.18)          & (x2.26)          &  & (x2.32)          & (x2.28)          & (x2.20)          & (x2.27)          &  & (x2.29)          & (x2.27)          & (x2.21)          & (x2.26)          \\
                                       & A                  & 2.76             & 2.73             & 2.73             & 2.74             &  & 2.77             & 2.73             & 2.73             & 2.74             &  & 2.77             & 2.73             & 2.74             & 2.75             \\
\midrule
\multirow{3}{*}{SAM+Recycle}           & \multirow{2}{*}{T} & 61.11            & 70.43            & 62.20            & 64.58            &  & 60.66            & 69.98            & 63.41            & 64.68            &  & 60.63            & 69.85            & 63.39            & 64.62            \\
                                       &                    & (x2.29)          & (x2.25)          & (x2.24)          & (x2.26)          &  & (x2.28)          & (x2.23)          & (x2.29)          & (x2.26)          &  & (x2.28)          & (x2.23)          & (x2.28)          & (x2.26)          \\
                                       & A                  & 2.71             & 2.73             & 2.68             & 2.71             &  & 2.69             & 2.74             & 2.69             & 2.71             &  & 2.68             & 2.71             & 2.67             & 2.69             \\
\midrule
\multirow{3}{*}{\textbf{STAND (Ours)}} & \multirow{2}{*}{T} & \textbf{64.99}   & \textbf{83.47}   & \textbf{69.70}   & \textbf{72.72}   &  & \textbf{66.88}   & \textbf{87.02}   & \textbf{71.83}   & \textbf{75.24}   &  & \textbf{69.15}   & \textbf{91.17}   & \textbf{74.14}   & \textbf{78.15}   \\
                                       &                    & \textbf{(x2.44)} & \textbf{(x2.66)} & \textbf{(x2.51)} & \textbf{(x2.55)} &  & \textbf{(x2.51)} & \textbf{(x2.78)} & \textbf{(x2.59)} & \textbf{(x2.63)} &  & \textbf{(x2.60)} & \textbf{(x2.91)} & \textbf{(x2.67)} & \textbf{(x2.74)} \\
                                       & A                  & \textbf{3.21}    & \textbf{3.48}    & \textbf{3.30}    & \textbf{3.33}    &  & \textbf{3.35}    & \textbf{3.70}    & \textbf{3.47}    & \textbf{3.51}    &  & \textbf{3.46}    & \textbf{3.90}    & \textbf{3.64}    & \textbf{3.67}    \\
\midrule
\multicolumn{16}{c}{\textit{\cellcolor[HTML]{EFEFEF}DeepSeek-R1-Distill-Qwen-14B}}                                                                                                                                                                                                                                            \\
\midrule
Plain                                  & T                  & 17.76            & 18.16            & 17.43            & 17.78            &  & 17.76            & 18.16            & 17.43            & 17.78            &  & 17.76            & 18.16            & 17.43            & 17.78            \\
\midrule
\multirow{3}{*}{Eagle-2}               & \multirow{2}{*}{T} & 25.38            & 24.86            & 21.89            & 24.04            &  & 25.38            & 24.86            & 21.89            & 24.04            &  & 25.38            & 24.86            & 21.89            & 24.04            \\
                                       &                    & (x1.43)          & (x1.37)          & (x1.26)          & (x1.35)          &  & (x1.43)          & (x1.37)          & (x1.26)          & (x1.35)          &  & (x1.43)          & (x1.37)          & (x1.26)          & (x1.35)          \\
                                       & A                  & 2.72             & 2.44             & 2.51             & 2.56             &  & 2.72             & 2.44             & 2.51             & 2.56             &  & 2.72             & 2.44             & 2.51             & 2.56             \\
\midrule
\multirow{3}{*}{PLD}                   & \multirow{2}{*}{T} & 24.37            & 26.6             & 23.36            & 24.78            &  & 25.44            & 27.36            & 23.96            & 25.59            &  & 26.35            & 28.43            & 24.97            & 26.58            \\
                                       &                    & (x1.37)          & (x1.46)          & (x1.34)          & (x1.39)          &  & (x1.43)          & (x1.51)          & (x1.37)          & (x1.44)          &  & (x1.48)          & (x1.57)          & (x1.43)          & (x1.49)          \\
                                       & A                  & 1.74             & 1.82             & 1.74             & 1.77             &  & 1.84             & 1.91             & 1.81             & 1.85             &  & 1.92             & 2.00                & 1.88             & 1.93             \\
\midrule
\multirow{3}{*}{ANPD}                  & \multirow{2}{*}{T} & 25.74            & 28.21            & 24.78            & 26.24            &  & 26.12            & 29.51            & 25.63            & 27.09            &  & 26.49            & 30.62            & 26.32            & 27.81            \\
                                       &                    & (x1.45)          & (x1.55)          & (x1.42)          & (x1.48)          &  & (x1.47)          & (x1.63)          & (x1.47)          & (x1.52)          &  & (x1.49)          & (x1.69)          & (x1.51)          & (x1.56)          \\
                                       & A                  & 1.87             & 1.97             & 1.87             & 1.90             &  & 1.91             & 2.04             & 1.93             & 1.96             &  & 1.96             & 2.13             & 1.99             & 2.03             \\
\midrule
\multirow{3}{*}{SAM}                   & \multirow{2}{*}{T} & 25.22            & 28.03            & 24.41            & 25.89            &  & 26.11            & 29.39            & 25.37            & 26.96            &  & 27.25            & 30.59            & 26.67            & 28.17            \\
                                       &                    & (x1.42)          & (x1.54)          & (x1.40)          & (x1.46)          &  & (x1.47)          & (x1.62)          & (x1.46)          & (x1.52)          &  & (x1.53)          & (x1.68)          & (x1.53)          & (x1.58)          \\
                                       & A                  & 1.78             & 1.85             & 1.79             & 1.81             &  & 1.88             & 1.95             & 1.87             & 1.90             &  & 1.98             & 2.06             & 1.96             & 2.00             \\
\midrule
\multirow{3}{*}{Recycle}               & \multirow{2}{*}{T} & 34.97            & 38.99            & 34.05            & 36.00            &  & 35.06            & 38.89            & 33.98            & 35.98            &  & 35.31            & 38.81            & 33.96            & 36.03            \\
                                       &                    & (x1.97)          & (x2.15)          & (x1.95)          & (x2.02)          &  & (x1.97)          & (x2.14)          & (x1.95)          & (x2.02)          &  & (x1.99)          & (x2.14)          & (x1.95)          & (x2.03)          \\
                                       & A                  & 2.78             & 2.73             & 2.72             & 2.74             &  & 2.77             & 2.73             & 2.72             & 2.74             &  & 2.77             & 2.74             & 2.72             & 2.74             \\
\midrule
\multirow{3}{*}{SAM+Recycle}           & \multirow{2}{*}{T} & 34.81            & 38.24            & 34.15            & 35.73            &  & 35.16            & 38.57            & 34.19            & 35.97            &  & 35.53            & 38.99            & 34.31            & 36.28            \\
                                       &                    & (x1.96)          & (x2.11)          & (x1.96)          & (x2.01)          &  & (x1.98)          & (x2.12)          & (x1.96)          & (x2.02)          &  & (x2.00)          & (x2.15)          & (x1.97)          & (x2.04)          \\
                                       & A                  & 2.70              & 2.71             & 2.65             & 2.69             &  & 2.71             & 2.71             & 2.66             & 2.69             &  & 2.72             & 2.71             & 2.65             & 2.69             \\
\midrule
\multirow{3}{*}{\textbf{STAND (Ours)}} & \multirow{2}{*}{T} & \textbf{37.56}   & \textbf{43.71}   & \textbf{38.71}   & \textbf{39.99}   &  & \textbf{39.13}   & \textbf{46.81}   & \textbf{40.45}   & \textbf{42.13}   &  & \textbf{40.76}   & \textbf{49.11}   & \textbf{42.72}   & \textbf{44.20}   \\
                                       &                    & \textbf{(x2.11)} & \textbf{(x2.41)} & \textbf{(x2.22)} & \textbf{(x2.25)} &  & \textbf{(x2.20)} & \textbf{(x2.58)} & \textbf{(x2.32)} & \textbf{(x2.37)} &  & \textbf{(x2.30)} & \textbf{(x2.70)} & \textbf{(x2.45)} & \textbf{(x2.49)} \\
                                       & A                  & \textbf{3.16}    & \textbf{3.42}    & \textbf{3.29}    & \textbf{3.29}    &  & \textbf{3.28}    & \textbf{3.63}    & \textbf{3.47}    & \textbf{3.46}    &  & \textbf{3.42}    & \textbf{3.86}    & \textbf{3.65}    & \textbf{3.64}   
\\ \bottomrule
\end{tabular}
}
\label{tab:main}
\end{table*}

In \Cref{fig:main} and \Cref{tab:main}, we evaluate \mname's performance in multi-trajectory inference, where we generate multiple candidate answers by sequentially producing k independent reasoning traces and then aggregate the results.

As shown in \Cref{fig:main}, \mname significantly improves decoding efficiency, achieving equivalent performance to plain decoding in less than 40\% the time. \Cref{tab:main} provides detailed throughput and acceptance length comparisons across methods. \mname not only achieves the highest throughput but also maintains longer acceptance lengths compared to baselines. Importantly, both metrics improve as we increase the number of trajectories, making \mname's speedup advantage more pronounced with increased compute scaling.

Notably, Token Recycle's performance remains flat despite increasing trajectories, unlike other model-free approaches. This limitation likely comes from its lookup table update strategy, which replaces rather than aggregates information from new trajectories. While this approach may offer
some drafting speed benefits, \mname's superior and scaling-dependent performance suggests that aggregating historical information is more beneficial than harmful for test-time scaling.

\subsection{Evaluation on single-trajectory decoding}
\label{sec:experiments-singletraj}

% CRv2-ACTIVE
\begin{table}[!h]
\centering
\small
\caption{\textbf{Single-trajectory evaluations.} We report the throughput (T) and acceptance length (A) for generating a single sequence with DeepSeek-R1-Distill-Qwen-7B and 14B. Best results are shown in \textbf{bold}.}
\label{tab:singletraj}
\adjustbox{width=\linewidth}{
\begin{tabular}{@{}lccccc@{}}
\toprule
                                       &                    & AIME             & GPQA             & LCB              & Avg.             \\
\midrule
\multicolumn{6}{c}{\textit{\cellcolor[HTML]{EFEFEF}DeepSeek-R1-Distill-Qwen-7B}}                                                                                \\
\midrule
Plain                                  & T                  & 26.63            & 31.34            & 27.75            & 28.57            \\
\midrule
\multirow{3}{*}{Eagle-2}               & \multirow{2}{*}{T} & 29.91            & 31.69            & 27.61            & 29.74            \\
                                       &                    & (x1.12)          & (x1.01)          & (x0.99)          & (x1.04)          \\
                                       & A                  & 2.21             & 1.99             & 2.13             & 2.11             \\
\midrule
\multirow{3}{*}{PLD}                   & \multirow{2}{*}{T} & 44.34            & 42.84            & 43.40            & 43.53            \\
                                       &                    & (x1.67)          & (x1.37)          & (x1.56)          & (x1.52)          \\
                                       & A                  & 1.72             & 1.64             & 1.59             & 1.65             \\
\midrule
\multirow{3}{*}{ANPD}                  & \multirow{2}{*}{T} & 46.18            & 54.05            & 44.79            & 48.34            \\
                                       &                    & (x1.73)          & (x1.72)          & (x1.61)          & (x1.69)          \\
                                       & A                  & 1.88             & 1.82             & 1.80             & 1.83             \\
\midrule
\multirow{3}{*}{SAM}                   & \multirow{2}{*}{T} & 40.85            & 48.45            & 42.92            & 44.07            \\
                                       &                    & (x1.53)          & (x1.55)          & (x1.55)          & (x1.54)          \\
                                       & A                  & 1.69             & 1.69             & 1.80             & 1.73             \\
\midrule
\multirow{3}{*}{Recycle}               & \multirow{2}{*}{T} & 60.61            & 71.00            & 60.12            & 63.91            \\
                                       &                    & (x2.28)          & (x2.27)          & (x2.17)          & (x2.24)          \\
                                       & A                  & 2.73             & 2.71             & 2.73             & 2.72             \\
\midrule
\multirow{3}{*}{SAM+Recycle}           & \multirow{2}{*}{T} & 61.15            & 71.51            & 62.78            & 65.15            \\
                                       &                    & (x2.30)          & (x2.28)          & (x2.26)          & (x2.28)          \\
                                       & A                  & 2.70             & 2.81             & 2.69             & 2.73             \\
\midrule
\multirow{3}{*}{\textbf{STAND (Ours)}} & \multirow{2}{*}{T} & \textbf{61.79}   & \textbf{75.39}   & \textbf{66.41}   & \textbf{67.86}   \\
                                       &                    & \textbf{(x2.32)} & \textbf{(x2.41)} & \textbf{(x2.39)} & \textbf{(x2.38)} \\
                                       & A                  & \textbf{3.07}    & \textbf{3.05}    & \textbf{3.01}    & \textbf{3.04}    \\
\midrule
\multicolumn{6}{c}{\textit{\cellcolor[HTML]{EFEFEF}DeepSeek-R1-Distill-Qwen-14B}}                                                                               \\
\midrule
Plain                                  & T                  & 17.76            & 18.16            & 17.43            & 17.78            \\
\midrule
\multirow{3}{*}{Eagle-2}               & \multirow{2}{*}{T} & 25.38            & 24.86            & 21.89            & 24.04            \\
                                       &                    & (x1.43)          & (x1.37)          & (x1.26)          & (x1.35)          \\
                                       & A                  & 2.72             & 2.44             & 2.51             & 2.56             \\
\midrule
\multirow{3}{*}{PLD}                   & \multirow{2}{*}{T} & 21.82            & 24.97            & 21.76            & 22.85            \\
                                       &                    & (x1.23)          & (x1.38)          & (x1.25)          & (x1.28)          \\
                                       & A                  & 1.61             & 1.64             & 1.58             & 1.61             \\
\midrule
\multirow{3}{*}{ANPD}                  & \multirow{2}{*}{T} & 25.60            & 26.40            & 23.16            & 25.05            \\
                                       &                    & (x1.44)          & (x1.45)          & (x1.33)          & (x1.41)          \\
                                       & A                  & 1.76             & 1.79             & 1.76             & 1.77             \\
\midrule
\multirow{3}{*}{SAM}                   & \multirow{2}{*}{T} & 23.26            & 25.38            & 22.36            & 23.67            \\
                                       &                    & (x1.31)          & (x1.40)          & (x1.28)          & (x1.33)          \\
                                       & A                  & 1.63             & 1.65             & 1.63             & 1.64             \\
\midrule
\multirow{3}{*}{Recycle}               & \multirow{2}{*}{T} & 33.71            & \textbf{38.91}            & 33.85            & 35.49            \\
                                       &                    & (x1.90)          & \textbf{(x2.14)}          & (x1.94)          & (x2.00)          \\
                                       & A                  & 2.77             & 2.73             & 2.71             & 2.74             \\
\midrule
\multirow{3}{*}{SAM+Recycle}           & \multirow{2}{*}{T} & 34.35            & 37.53            & 34.45            & 35.44            \\
                                       &                    & (x1.93)          & (x2.07)          & (x1.98)          & (x1.99)          \\
                                       & A                  & 2.67             & 2.72             & 2.70             & 2.70             \\
\midrule
\multirow{3}{*}{\textbf{STAND (Ours)}} & \multirow{2}{*}{T} & \textbf{34.52}   & 38.71   & \textbf{34.86}   & \textbf{36.03}   \\
                                       &                    & \textbf{(x1.94)} & (x2.13) & \textbf{(x2.00)} & \textbf{(x2.03)} \\
                                       & A                  & \textbf{2.91}    & \textbf{3.00}    & \textbf{2.93}    & \textbf{2.95}   
\\ \bottomrule
\end{tabular}
}
\end{table}

While \mname is primarily designed to leverage information across multiple reasoning trajectories, we also evaluate its performance on single-trajectory generation, where the model only produces one long reasoning chain. As shown in \Cref{tab:singletraj}, \mname achieves both the highest acceptance length and throughput in most scenarios, demonstrating its effectiveness even when generating individual solutions.

\subsection{Evaluation on diverse inference patterns}
\label{sec:experiments-batch-dvts}

% CRv2-ACTIVE
\begin{table}[!h]
\centering
\small
\caption{\textbf{Batch decoding evaluations.} We report the throughput (T) and acceptance length (A) for batch decoding with DeepSeek-R1-Distill-Qwen-7B. Both speculative decoding approaches use trees optimized with OpenThoughts-114k. Best results are shown in \textbf{bold}.}
\label{tab:batch}
\adjustbox{width=\linewidth}{
\begin{tabular}{@{}lccccc@{}}
\toprule
                                       &                    & AIME             & GPQA             & LCB              & Avg.             \\
\midrule
\multicolumn{6}{c}{\textit{\cellcolor[HTML]{EFEFEF}Batch Size 4}}                                                                                                 \\
\midrule
Plain                                  & T                    & 89.04            & 88.32            & 92.64            & 90.00            \\
\midrule
\multirow{3}{*}{Recycle}               & \multirow{2}{*}{T}   & 90.51            & 98.41            & 89.55            & 92.82            \\
                                       &                      & (x1.02)          & (x1.11)          & (x0.97)          & (x1.03)          \\
                                       & A                    & 1.77             & 1.83             & 1.72             & 1.77             \\
\midrule
\multirow{3}{*}{\textbf{STAND (Ours)}} & \multirow{2}{*}{T}   & \textbf{127.58}  & \textbf{134.64}  & \textbf{122.09}  & \textbf{128.10}  \\
                                       &                      & \textbf{(x1.43)} & \textbf{(x1.52)} & \textbf{(x1.32)} & \textbf{(x1.42)} \\
                                       & A                    & \textbf{2.57}    & \textbf{2.74}    & \textbf{2.57}    & \textbf{2.63}    \\
\midrule
\multicolumn{6}{c}{\textit{\cellcolor[HTML]{EFEFEF}Batch Size 8}}                                                                                                 \\
\midrule
Plain                                  & T                    & 111.62           & 106.23           & 114.73           & 110.86           \\
\midrule
\multirow{3}{*}{Recycle}               & \multirow{2}{*}{T}   & 99.18            & 105.69           & 99.2             & 101.36           \\
                                       &                      & (x0.89)          & (x0.99)          & (x0.86)          & (x0.91)          \\
                                       & A                    & 1.66             & 1.75             & 1.63             & 1.68             \\
\midrule
\multirow{3}{*}{\textbf{STAND (Ours)}} & \multirow{2}{*}{T}   & \textbf{148.08}  & \textbf{154.68}  & \textbf{149.41}  & \textbf{150.72}  \\
                                       &                      & \textbf{(x1.33)} & \textbf{(x1.46)} & \textbf{(x1.30)} & \textbf{(x1.36)} \\
                                       & A                    & \textbf{2.70}     & \textbf{2.86}    & \textbf{2.69}    & \textbf{2.75}   
\\
\bottomrule
\end{tabular}
}
\end{table}

% CRv2-ACTIVE
\begin{table}[!h]
\centering
\small
\caption{\textbf{Test-time tree search evaluations.} We report the throughput (T) and acceptance length (A) for performing Diverse Verifier Tree Search (DVTS) with DeepSeek-R1-Distill-Qwen-7B. Both speculative decoding approaches use trees optimized with OpenThoughts-114k. Best results are shown in \textbf{bold}.}
\label{tab:dvts}
\adjustbox{width=\linewidth}{
\begin{tabular}{@{}lccccc@{}}
\toprule
                                       &                    & AIME             & GPQA             & LCB              & Avg.             \\
\midrule
Plain                                  & T                    & 33.35            & 33.12            & 32.22            & 32.90            \\
\midrule
\multirow{3}{*}{Recycle}               & \multirow{2}{*}{T}   & 71.52            & 70.64            & 69.47            & 70.54            \\
                                       &                      & (x2.14)          & (x2.13)          & (x2.16)          & (x2.14)          \\
                                       & A                    & 2.73             & 2.69             & 2.74             & 2.72             \\
\midrule
\multirow{3}{*}{\textbf{STAND (Ours)}} & \multirow{2}{*}{T}   & \textbf{82.62}   & \textbf{86.93}   & \textbf{80.97}   & \textbf{83.51}   \\
                                       &                      & \textbf{(x2.48)} & \textbf{(x2.62)} & \textbf{(x2.51)} & \textbf{(x2.54)} \\
                                       & A                    & \textbf{3.55}    & \textbf{3.71}    & \textbf{3.51}    & \textbf{3.59}   
\\ \bottomrule
\end{tabular}
}
\end{table}

We extend our evaluation to more diverse and practical inference patterns, focusing on batch decoding and test-time tree search. In the batch decoding setup, multiple reasoning traces are generated simultaneously, and the N-gram drafters can only leverage the content produced up to the current decoding step. As shown in \Cref{tab:batch}, \mname consistently achieves significant speedups even with a batch size of 8, whereas Token Recycle provides only marginal improvements at batch size 4 and even degrades performance at batch size 8.

Test-time tree search represents another widely used pattern for scaling inference, where the model generates multiple candidate reasoning steps and dynamically selects among them. In \Cref{tab:dvts}, we evaluate \mname within the Diverse Verifier Tree Search (DVTS, \cite{dvts})  framework. As the results show, \mname consistently outperforms Token Recycle, achieving an average speedup of 2.54$\times$.

\subsection{Ablations and analysis}
\label{sec:experiments-ablations}

We evaluate key components of \mname through ablation studies and further analysis. Our ablation studies examine the impact of stochastic drafting and the Gumbel-Top-K optimization trick, followed by an investigation of our tree optimization approach. We then analyze the structural characteristics of the optimized trees to better understand the patterns that emerge from our method.

% CRv2-ACTIVE
\begin{table}[t]
\centering
\small
\caption{\textbf{Effect of Stochastic Drafting.} We report the throughput (T) and acceptance length (A) for generating 4 sequences with DeepSeek-R1-Distill-Qwen-7B.}
\label{tab:sampling}
\adjustbox{width=\linewidth}{
\begin{tabular}{@{}lccccc@{}}
\toprule
                                & \multicolumn{1}{l}{} & AIME    & GPQA    & LCB     & Avg.    \\
\midrule
Plain                           & T                    & 26.63   & 31.34   & 27.75   & 28.57   \\
\midrule
\multirow{3}{*}{Deterministic}  & \multirow{2}{*}{T}   & 62.13   & 73.67   & 63.44   & 66.41   \\
                                &                      & (x2.33) & (x2.35) & (x2.29) & (x2.32) \\
                                & A                    & 2.94    & 2.98    & 2.90    & 2.94    \\
\midrule
\multirow{3}{*}{Stochastic}     & \multirow{2}{*}{T}   & 63.44   & 81.20   & 65.90   & 70.18   \\
                                &                      & (x2.38) & (x2.59) & (x2.37) & (x2.46) \\
                                & A                    & 3.24    & 3.56    & 3.29    & 3.36    \\
\midrule
\multirow{3}{*}{+ Gumbel-Top-K} & \multirow{2}{*}{T}   & 64.99   & 83.47   & 69.70   & 72.72   \\
                                &                      & (x2.44) & (x2.66) & (x2.51) & (x2.55) \\
                                & A                    & 3.21    & 3.48    & 3.30    & 3.33   
\\ \bottomrule
\end{tabular}
}
\end{table}

\paragraph{Effect of stochastic drafting.}
In \Cref{tab:sampling}, we compare three drafting approaches: deterministic drafting, basic stochastic drafting (using PyTorch's multinomial sampling), and our optimized stochastic drafting with Gumbel-Top-K. For fair comparison, we separately perform tree optimization for determinisic drafting and stochastic drafting. Stochastic drafting consistently achieves higher acceptance lengths across all tasks, resulting in improved throughput compared to deterministic drafting. Our Gumbel-Top-K optimization further improves performance by maintaining similar acceptance lengths while significantly reducing latency, leading to even higher throughput.

% CRv2-ACTIVE
\begin{table}[t]
\centering
\small
\caption{\textbf{Effect of tree optimization.} Comparison of throughput and acceptance length when generating 4 sequences with DeepSeek-R1-Distill-Qwen-7B on two datasets: AIME-2024 and GPQA-Diamond. We compare two types of static trees: the heuristic trees from Token Recycle and our data-optimized trees, optimized on AIME-2024.
Best results are shown in \textbf{bold}.}
\label{tab:treeopt}
\adjustbox{width=\linewidth}{
\begin{tabular}{@{}lcccc@{}}
\toprule
           & \multicolumn{2}{c}{AIME} & \multicolumn{2}{c}{GPQA (OOD)} \\
           & Heuristic      & Optimized     & Heuristic      & Optimized     \\
\midrule
Throughput & 59.96          & \textbf{64.99}         & 77.32          & \textbf{83.47}         \\
Acc. Lens  & 3.17           & \textbf{3.21}          & 3.35           & \textbf{3.48} 
    
\\
\bottomrule
\end{tabular}
}
\end{table}

\paragraph{Effect of tree optimization.} In \Cref{tab:treeopt}, we showcase the effectiveness of our tree optimization technique. We compare the performance of a heuristic tree originally used by Token Recycle \citep{luo2024turning} with our tree, optimized on the AIME-2024 dataset. The results demonstrates that the optimized tree improves performance on both AIME-2024 and GPQA-Diamond, showcasing that the optimization not only works within the same dataset, but also generatlizes to out-of-domain (OOD) tasks.

% CRv2-ACTIVE
\begin{figure}[h]
\centering
\includegraphics[width=\linewidth]{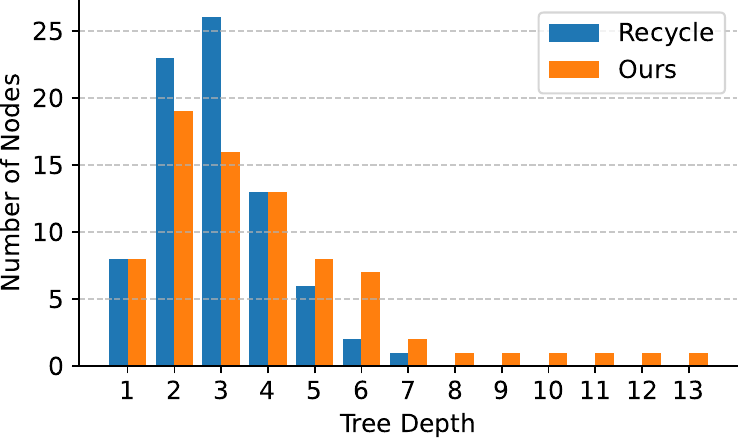}
\caption{\textbf{Structure of the Optimized Tree.} We report the number of nodes at specific tree depths for draft trees optimized for each Token Recycle and \mname. Both trees are optimized on AIME-2024 dataset with DeepSeek-R1-Distill-Qwen-7B.}
\label{fig:tree_structure}
\end{figure}

\paragraph{Tree structure analysis.} We analyze how different drafting approaches lead to different optimal tree structures by comparing trees optimized for \mname versus Token Recycle. As shown in \Cref{fig:tree_structure}, the tree optimized for \mname reaches greater depths, extending to 13 levels compared to 7 levels in the Token Recycle-optimized tree. This difference likely stems from \mname's higher acceptance rate, which favors deeper, narrower tree structures under the same tree size budget.

A distinctive feature of \mname's optimized tree is its long tail structure, with single nodes at depths 8 through 13. This pattern suggests the presence of occasional long, deterministic sequences, possibly arising from consistent patterns found across multiple reasoning trajectories.
% CRv2-ACTIVE
\section{Conclusion}
\label{sec:conclusion}

In this work, we introduced \mname, a model-free speculative decoding approach that accelerates language model reasoning while maintaining accuracy. By utilizing reasoning trajectory redundancy and historical logit information, \mname significantly improves throughput over standard auto-regressive decoding and existing alternatives, offering an efficient solution for scaling AI reasoning systems.

\clearpage

\section*{Limitations}

While \mname demonstrates strong performance across diverse inference patterns, all measurements were conducted using the HuggingFace implementation, which is less optimized than popular serving frameworks such as vLLM or SGLang. Although we expect the benefits of \mname to extend to these optimized frameworks, this has not yet been verified. In addition, the current N-gram lookup operation is implemented in Python, which may introduce a slowdown. Latency could be further reduced with a more optimized implementation of the N-gram module.

\section*{Acknowledgements}

For WS and JS: This work was supported by Institute for Information \& communications Technology Promotion(IITP) grant funded by the Korea government(MSIT) (No.RS-2019-II190075 Artificial Intelligence Graduate School Program (KAIST); No. RS-2024-00509279, Global AI Frontier Lab). This work was also supported by the NIPA(National IT Industry Promotion Agency), through the Ministry of Science and ICT (Hyperscale AI flagship project).

\bibliography{custom}

% CRv2-ACTIVE
\clearpage

\appendix

\section{Experimental Details}

\subsection{Initial tree for optimization}
\label{sec:app-bigtree}

% CRv2-ACTIVE
\begin{figure}[h]
\centering
\includegraphics[width=\linewidth]{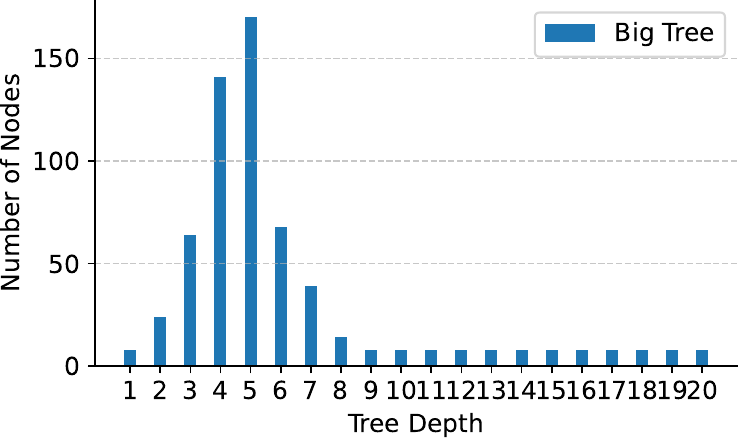}
\caption{\textbf{Structure of the initial tree.} We report the number of nodes at specific tree depths for the initial tree used for tree optimization.}
\label{fig:big_tree_structure}
\end{figure}

To initialize data-driven tree optimization, we heuristically applied predefined rules that assign the number of child nodes based on node depth and position. In particular, leftmost nodes at each level receive more children, as they are more likely to yield higher acceptance probabilities. The pseudocode for this initialization is shown in \Cref{alg:init-tree}.

As illustrated in \Cref{fig:big_tree_structure}, the resulting tree has a maximum depth of 20 and 625 total nodes.

\subsection{Experiment Details}

For the batch decoding experiments, we modified the single-batch inference code to sequentially construct draft trees for all sequences in a batch, and then verify them in parallel. This design choice was made because, unlike model-based drafters that benefit substantially from forwarding multiple samples in parallel (even during the drafting stage), N-gram–based drafters inherently rely on sequential memory lookups. For the DVTS experiments, we adopted the self-evaluation strategy from SpecReason \citep{pan2025specreason} to determine which reasoning step to accept.

\section{Further discussions}

\subsection{Effect of tree optimization dataset}

In this section, we analyze the impact of the dataset used for tree optimization. To evaluate this effect, we optimized the draft tree on the OpenThoughts-114k dataset and then evaluated STAND on AIME-2024, rather than optimizing directly on AIME-2024. As shown in \Cref{tab:app-dataset}, STAND maintains high throughput and acceptance length under both configurations, in some cases even outperforming the tree optimized on AIME-2024. These results suggest that STAND is robust to the choice of initialization dataset.

\subsection{Remarks on Eagle-2 performance}

% CRv2-ACTIVE
\begin{table}[!h]
\centering
\small
\caption{\textbf{Eagle-2 acceptance lengths.} We report the acceptance lengths of the Eagle-2 model across different context lengths, using DeepSeek-R1-Distill-Qwen-14B.}
\label{tab:app-eagle2}
\adjustbox{width=\linewidth}{
\begin{tabular}{@{}lccccc@{}}
\toprule
Input Length  & 0-2k & 2k-4k & 4k-8k & 8k-16k & 16k-32k \\
\midrule
AIME-2024     & 2.89 & 2.79  & 2.68  & 2.61   & 2.47    \\
GPQA-Diamond  & 2.60 & 2.50  & 2.38  & 2.31   & 2.28    \\
LiveCodeBench & 2.76 & 2.66  & 2.50  & 2.38   & 2.14   
\\ \bottomrule
\end{tabular}
}
\end{table}

Somewhat surprisingly, we observed lower acceptance lengths for Eagle-2 compared to those originally reported in the paper. One potential explanation is the long-context setup used in our evaluation. Whereas most prior Eagle-2 benchmarks focused on short inputs (fewer than 2k tokens), our experiments trained Eagle drafters to handle up to 32k tokens in order to support long-form reasoning. This broader context window may reduce acceptance length as a trade-off.

We also observed that acceptance lengths tend to degrade with longer inputs, as shown in \Cref{tab:app-eagle2}, which provides another possible explanation for the lower average acceptance lengths in our experiments.

\subsection{Performance on non-reasoning model}

% CRv2-ACTIVE
\begin{table}[ht]
\centering
\small
\caption{\textbf{Non-reasoning model evaluation.} We report the throughput (T) and acceptance length (A) for generating multiple sequences with Qwen2.5-7B-Instruct with batch decoding.}
\label{tab:app-qwen}
\adjustbox{width=\linewidth}{
\begin{tabular}{@{}lcccc@{}}
\toprule
 & Batch Size & AIME (T / A) & GPQA (T / A) \\
\midrule
Plain        & 4 & 96.64 / -- & 130.00 / -- \\
Recycle      & 4 & 115.75 / 2.06 & 133.53 / 1.99 \\
\textbf{STAND (Ours)} & 4 & \textbf{152.59} / \textbf{2.65} & \textbf{152.07} / \textbf{2.39} \\
\midrule
Plain        & 8 & 119.32 / -- & 166.05 / -- \\
Recycle      & 8 & 134.40 / 1.95 & 155.41 / 1.94 \\
\textbf{STAND (Ours)} & 8 & \textbf{187.43} / \textbf{2.82} & \textbf{198.09} / \textbf{2.62} \\
\bottomrule
\end{tabular}
}
\end{table}

While our analysis in the main text focused on reasoning models, namely the DeepSeek-R1-Distill-Qwen family, \mname can also be applied to test-time scaling with non-reasoning models. As shown in \Cref{tab:app-qwen}, \mname significantly outperforms Token Recycle in both throughput and acceptance length at batch sizes 4 and 8 with Qwen2.5-7B-Instruct.

% CRv2-ACTIVE
\begin{algorithm*}[!h]
\caption{Initial Tree Construction}
\label{alg:init-tree}
\begin{algorithmic}[1]
\State Initialize $\texttt{active\_list} \gets \{\texttt{root\_node}\}$
\State Initialize $\texttt{children\_list} \gets \emptyset$
\State $\texttt{num\_nodes} \gets 1$

\For{$\texttt{depth} = 0$ \textbf{to} $19$}
    \For{each $\texttt{node}$ in $\texttt{active\_list}$ with index $\texttt{i}$}
        \If{$\texttt{node.depth} = 0$}
            \State $\texttt{n\_children} \gets 8$
        \ElsIf{$\texttt{node.depth} = 1$}
            \State $\texttt{n\_children} \gets \max(8 - 2 \cdot \texttt{node.order}, 1)$
        \ElsIf{$\texttt{node.depth} = 2$}
            \State $\texttt{n\_children} \gets \max\!\Big(\lceil \tfrac{|\texttt{node.parent.children}|-1}{\texttt{node.order} \cdot 0.7 + 1}\rceil, 2\Big)$
        \ElsIf{$\texttt{node.depth} = 3$}
            \State $\texttt{n\_children} \gets \max\!\Big(\lceil \tfrac{|\texttt{node.parent.children}|-1}{\texttt{node.order} \cdot 0.7 + 1}\rceil, 2\Big)$
        \Else
            \State $\texttt{n\_children} \gets \max\!\Big(\lceil \tfrac{|\texttt{node.parent.children}|-1}{\texttt{node.order} \cdot 0.7 + 1}\rceil, 0\Big)$
        \EndIf

        \If{$\texttt{i} = 0$}
            \State $\texttt{n\_children} \gets \max(\texttt{n\_children}, 3)$
        \EndIf

        \For{$\texttt{i\_child} = 0$ \textbf{to} $\texttt{n\_children}-1$}
            \State Create new $\texttt{child\_node}$ with:
            \State \quad $\texttt{id} = \texttt{num\_nodes},$ 
            \State \quad $\texttt{depth} = \texttt{node.depth}+1,$ 
            \State \quad $\texttt{order} = \texttt{i\_child},$ 
            \State \quad $\texttt{parent} = \texttt{node}$
            \State $\texttt{num\_nodes} \gets \texttt{num\_nodes} + 1$
            \State Append $\texttt{child\_node}$ to $\texttt{node.children}$ and $\texttt{children\_list}$
        \EndFor
    \EndFor

    \State $\texttt{active\_list} \gets \texttt{children\_list}$
    \State $\texttt{children\_list} \gets \emptyset$
\EndFor
\end{algorithmic}
\end{algorithm*}
% CRv2-ACTIVE
\begin{table*}[!h]
\centering
\small
\caption{\textbf{Effect of tree optimization dataset.} We report the average throughput (T) and acceptance length (A) for the best baseline and \mname with trees optimized from AIME-2025 and OpenThoughts-114k. We evaluate each model on AIME-2024.}
\label{tab:app-dataset}
\begin{tabular}{lcccccccc}
\toprule
                           & \multicolumn{2}{c}{Single Trajectory} & \multicolumn{2}{c}{4 Trajectories} & \multicolumn{2}{c}{8 Trajectories} & \multicolumn{2}{c}{16 Trajectories} \\
\cmidrule(lr){2-3} \cmidrule(lr){4-5} \cmidrule(lr){6-7} \cmidrule(lr){8-9}
                           & T                  & A                & T                & A               & T                & A               & T                 & A               \\
\midrule
\multicolumn{9}{c}{\textit{\cellcolor[HTML]{EFEFEF}DeepSeek-R1-Distill-Qwen-7B}}                                                                                                                           \\
\midrule
Best Baseline              & 61.15              & 2.73             & 61.38            & 2.76            & 61.70            & 2.77            & 60.86             & 2.77            \\
STAND w/ AIME Tree         & 61.79              & 3.07             & 64.99            & 3.21            & 66.88            & 3.35            & 69.15             & 3.46            \\
STAND w/ OpenThoughts Tree & 62.75              & 3.04             & 65.33            & 3.24            & 68.58            & 3.35            & 70.63             & 3.49            \\
\midrule
\multicolumn{9}{c}{\textit{\cellcolor[HTML]{EFEFEF}DeepSeek-R1-Distill-Qwen-14B}}                                                                                                                          \\
\midrule
Best Baseline              & 34.35              & 2.77             & 34.97            & 2.78            & 35.16            & 2.71            & 35.53             & 2.72            \\
STAND w/ AIME Tree         & 34.52              & 2.91             & 37.56            & 3.16            & 39.13            & 3.28            & 40.76             & 3.42            \\
STAND w/ OpenThoughts Tree & 35.26              & 2.87             & 37.41            & 3.10            & 39.22            & 3.26            & 41.38             & 3.41           
\\ \bottomrule
\end{tabular}
\end{table*}

\onecolumn
\clearpage
\twocolumn

\subsection{Effect of tree depth on speed-up}
\label{app:tree-depth}

Prior work on model-based speculative decoding has investigated the trade-off between tree depth and speed-up. For example, POSS \citep{huang2025poss} reports that deeper draft trees can increase acceptance length, but at the cost of higher drafting latency, since each additional depth requires an extra forward pass of the draft model. This trade-off directly affects the overall speed-up ratio in model-based methods such as Eagle-2.

By contrast, STAND uses a static, pre-computed tree and performs efficient dictionary lookups rather than model forward passes. As it sequentially performs the dictionary lookup for each tree node, the drafting time does not necessarily increase with the tree depth. Even if lookup times increase slightly, the cost remains negligible compared to model-based approaches, thanks to the efficiency of dictionary lookups.

\begin{table}[]
\begin{tabular}{lccccc}
\toprule
Pruning Ratio & 0\% & 5\% & 10\% & 15\% & 20\% \\
\midrule
Initial & 0.8795 & 0.8781 & 0.8788 & 0.8791 & 0.8791 \\
Think-Attn & 0.8795 & 0.8744 & 0.8747 & 0.8745 & 0.8741
\\ \bottomrule
\end{tabular}
\end{table}
\end{document}